\documentclass[conference]{IEEEtran}
\usepackage{times}

\usepackage[numbers]{natbib}
\usepackage{multicol}
\usepackage[bookmarks=true,hidelinks]{hyperref}
\usepackage{amsmath}
\usepackage[dvipsnames]{xcolor}
\usepackage{graphicx}

\usepackage{amssymb}
\usepackage{graphicx}
\usepackage{subcaption}
\usepackage{multirow}
\usepackage{booktabs}
\usepackage{derivative}
\usepackage{mathtools}
\usepackage{algorithm}
\usepackage{algpseudocode}
\usepackage{flushend}
\usepackage{mdframed}
\usepackage[normalem]{ulem}

\DeclareMathOperator*{\argmin}{arg\,min}
\newcommand{\ph}[1]{{\textbf{#1}:}} 

\begin{document}

\begin{titlepage}

        This paper has been published in the proceedings of Robotics: Science and Systems, 2024. Please cite this paper as:

    \vspace{1em}

    \begin{verbatim}
    @inproceedings{Ginting2024Seek,
        title = "{SEEK}: Semantic Reasoning for Object Goal Navigation 
                 in Real World Inspection Tasks",
        author = {Muhammad Fadhil Ginting and Sung-Kyun Kim and  David D. Fan and 
                  Matteo Palieri and Mykel J. Kochenderfer and Ali-akbar Agha-mohammadi},
        year = {2024},
        booktitle = "Proc. of Robotics: Science and Systems",
    }
    \end{verbatim}
    
    \vfill 
\end{titlepage}

\title{SEEK: Semantic Reasoning for Object Goal Navigation in Real World Inspection Tasks}

\author{\authorblockN{Muhammad Fadhil Ginting$^{1}$, Sung-Kyun Kim$^{2}$, David D. Fan$^{2}$, Matteo Palieri$^{2}$, \\
Mykel J. Kochenderfer$^{1}$, and Ali-akbar Agha-mohammadi$^{2}$}
\authorblockA{$^{1}$Stanford University, $^{2}$Field AI
}
}

\maketitle

\begin{abstract}
This paper addresses the problem of object-goal navigation in autonomous inspections in real-world environments. 
Object-goal navigation is crucial to enable effective inspections in various settings, often requiring the robot to identify the target object within a large search space. 
Current object inspection methods fall short of human efficiency because they typically cannot bootstrap prior and common sense knowledge as humans do.
In this paper, we introduce a framework that enables robots to use semantic knowledge from prior spatial configurations of the environment and semantic common sense knowledge. 
We propose SEEK (Semantic Reasoning for Object Inspection Tasks) that combines semantic prior knowledge with the robot's observations to search for and navigate toward target objects more efficiently. 
SEEK maintains two representations: a Dynamic Scene Graph (DSG) and a Relational Semantic Network (RSN). 
The RSN is a compact and practical model that estimates the probability of finding the target object across spatial elements in the DSG.
We propose a novel probabilistic planning framework to search for the object using relational semantic knowledge. 
Our simulation analyses demonstrate that SEEK outperforms the classical planning and Large Language Models (LLMs)-based methods that are examined in this study in terms of efficiency for object-goal inspection tasks. 
We validated our approach on a physical legged robot in urban environments, showcasing its practicality and effectiveness in real-world inspection scenarios.

\end{abstract}
\IEEEpeerreviewmaketitle

\section{Introduction}
\begin{figure}[t]
    \centering
    \includegraphics[width=0.48\textwidth]{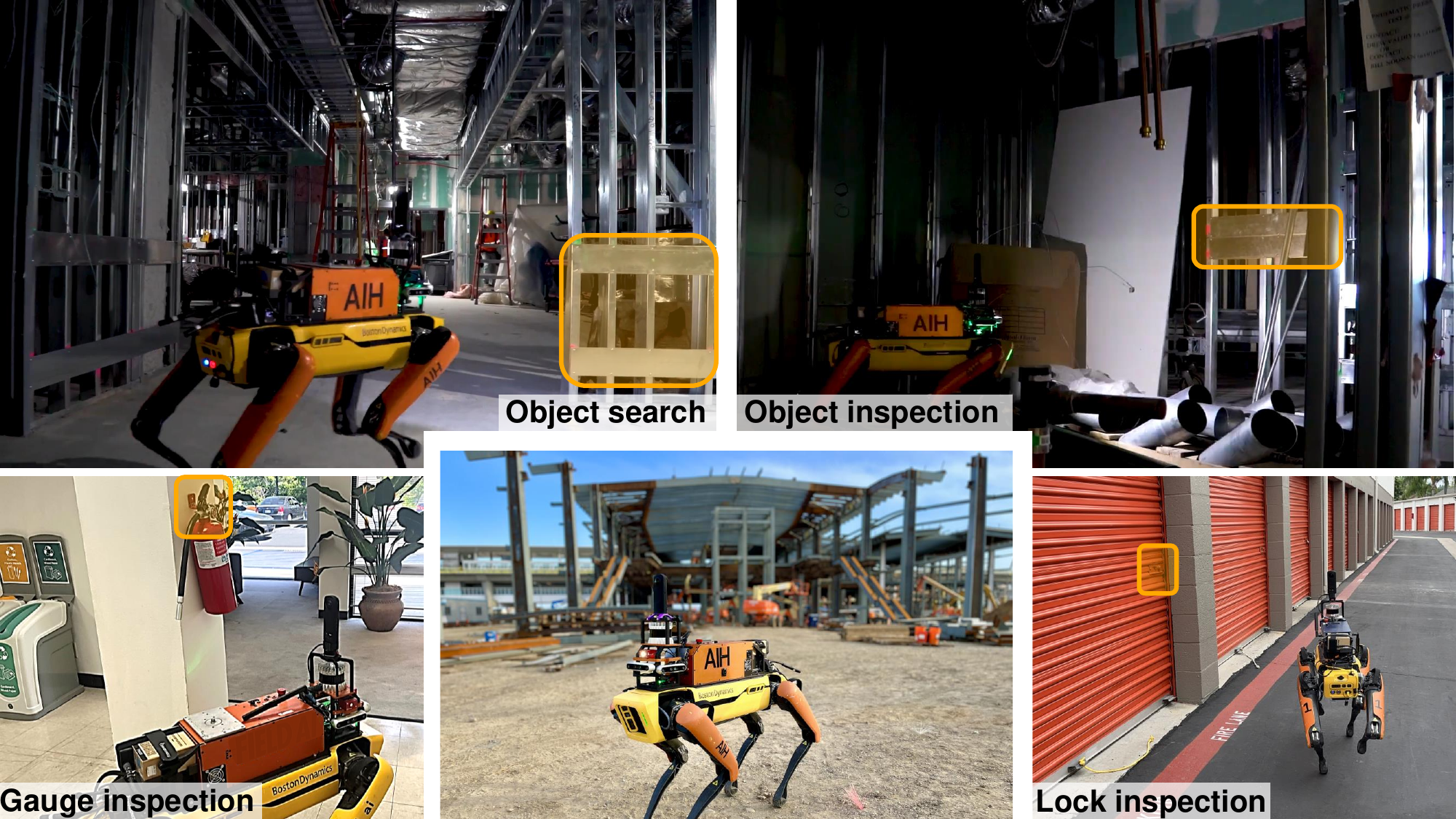}
    \caption{Autonomous object inspections in urban buildings and construction sites. The robot needs to search and inspect the target object. In this paper, we proposed a method to guide the robot to find the object using relational semantic knowledge.}
    \label{fig:cover_figure}
\end{figure}

Consider an autonomous robot searching and inspecting a target object in an unseen environment. 
For example, to inspect a fire extinguisher in a building, the robot needs to reason where it can find the object in the environment and search for the object efficiently. 
This task, known as an object-goal navigation problem, is one of the central problems in embodied AI research~\cite{batra2020objectnav}. 
This capability is crucial in a wide range of real-world applications, including urban inspection~\cite{tan2021automatic,bouman2020autonomous,lattanzi2017review}, home robots~\cite{wu2023tidybot}, monitoring of oil and gas sites~\cite{gehring2021anymal}, and exploration of subterranean environments~\cite{agha2021nebula, tranzatto2022cerberus, hudson2022heterogeneous}. 
This work addresses the object-goal navigation problem for autonomous inspection in unseen real-world environments~(\autoref{fig:cover_figure}).

Performing object-goal navigation in unseen environments presents three main challenges. 
The first challenge is the large search space for the robot to consider when deciding where to go and inspect. The target object may be located in one room among many rooms in a building, and thus, uninformed brute-force search can be significantly inefficient. 
The second challenge is the robot's inherent inability to use experience and common sense knowledge. 
Unlike humans, who draw upon prior knowledge and contextual understanding to refine their search strategies, the current state-of-the-practice object search methods are confined to sensory information onboard the robot. 
The third challenge is detecting the object under perceptual uncertainty due to the limited sensing range, inter-object occlusion, and detection and localization errors of the perception model on the robot. 
For example, planning for object search without probabilistic reasoning of the object detection accuracy could lead to inefficient search policy when there are false positives of the object detection~\cite{ginting2024semantic, yokoyama2024vlfm}. 
Decision making without probabilistic reasoning of prior semantic knowledge and perceptual uncertainty can lead to a myopic greedy policy or an overly conservative inefficient policy. 
These challenges underscore a fundamental barrier in object-goal navigation, highlighting the need for a new framework that mimics human reasoning capabilities.

To address these challenges, we develop a framework that uses prior environment information and semantic knowledge in a probabilistic manner.
While standard object-goal navigation problems only assume access to robot sensory readings~\cite{batra2020objectnav}, our approach can improve robot search efficiency by leveraging both \textit{i) prior spatial configuration} and \textit{ii) relational semantic knowledge} that are commonly accessible to humans performing inspection tasks. 
The environment spatial configuration, such as room layouts, floor plans, blueprints, and Building Information Modeling (BIM)~\cite{volk2014building}, provides a structured context for the robot to search the environment hierarchically. 
Additionally, relational semantic knowledge, such as typical locations of objects within the environment, can guide the object search.
For instance, knowing that fire extinguishers are often found near entrances or kitchens, we can direct the robot to search those regions to locate the object.
The advent of Large Language Models (LLMs) and language embeddings ~\cite{brown2020language,devlin2019bert} allows us to access this semantic knowledge without human assistance.
By leveraging both spatial and semantic information, we can formulate a \textit{probabilistic optimization} problem for more efficient and robust object-goal navigation under uncertainty. We solve for a non-myopic policy that prioritizes the region with a higher probability of containing the target object as well as takes into account the chance of not finding the target object in the selected region, its associated costs, and the best follow-up policy in a recursive fashion. 
This capability can set a new standard for object-goal navigation in real-world inspections.

In this paper, we propose SEEK (Semantic Reasoning for Object Inspection Tasks), a framework for object-goal navigation that pushes the state-of-the-practice of real-world inspections. 
SEEK maintains two environment representations: Dynamic Scene Graph (DSG) and a novel Relational Semantic Network (RSN). 
The DSG is a hierarchical world representation generated from a prior spatial configuration of the environment. 
The RSN encodes relational semantic knowledge between objects and the regions or rooms in the environment. 
We then solve the object-goal navigation problem using a novel probabilistic planning framework with relational semantic knowledge. 
Using DSG and RSN, we first formulate the global planning problem as a Markov decision process (MDP)~\cite{kochenderfer2022algorithms}. 
The computed global planning policy directs the robot to visit a room or perform local searches. 
We then use a finite state local controller to execute the global planning policy and search for the target object. 
This approach enables a principled way to probabilistically combine prior and common sense knowledge with the robot's observation. 
In contrast with current state-of-the-art methods, which require resource-intensive models~\cite{dorbala2023can} or large amount of data~\cite{chaplot2020object}, 
our RSN is a lightweight model trained with a small amount of data distilled from LLMs. 
This compact model size is crucial for real-world inspection robots, which often have limited computational resources and restricted internet access. 
The RSN updates its semantic knowledge over time based on the robot's observations in the specific environment.

To demonstrate the efficacy of SEEK, we evaluate our approach extensively in simulation and real-world inspections. 
First, we evaluate the efficiency of our approach against different classical and LLM-based planners in simulation using the Matterport3D datasets~\cite{Matterport3D} in the Habitat simulator~\cite{savva2019habitat}. 
Next, we compared our approach with an existing semantic-based inspection approach~\cite{ginting2024semantic} in the Gazebo simulator and showed that our approach significantly improved the object search efficiency. 
Finally, we demonstrate our approach in real-world autonomous inspection scenarios in an office building.

In summary, our technical contributions are as follows:
\begin{enumerate}
    \item We introduce SEEK, a framework that uses prior spatial configuration and relational semantic knowledge for semantic-guided navigation;    
    \item We propose a method to build a relational semantic network (RSN) using common-sense knowledge contained in an LLM;
    \item We design a probabilistic planning algorithm for object-goal navigation with relational semantic knowledge;
    \item We validate our framework through simulation and real-world demonstrations with a legged robot. 
\end{enumerate}

\section{Related Works}
\vspace{-3pt}
\ph{Planning for autonomous inspection}
Our approach is related to planning methods for autonomous robot exploration and inspection in real-world environments. 
Coverage planning is a well-studied method to explore and map environments efficiently~\cite{bouman2022adaptive, peltzer2022fig, cao2021tare, charrow2015information, bircher2015structural}. 
In addition to map the whole environment, inspection tasks often require the robot to closely examine specific target objects of interest in some part of the environment~\cite{hutter2018towards}. 
Current state-of-the-practice approaches to inspecting specific targets usually rely on predefining routes and observation points for the robot or placing identifiable tags, such as Apriltags~\cite{wang2016apriltag} or QR codes~\cite{BostonDynamics2023GraphNav}, making the process labor-intensive for humans. 
Recent approaches to addressing this problem involve training an object detector and incorporating the detection model into the planning process~\cite{dang2020autonomous,grinvald2019volumetric,dharmadhikari2023semantics,ginting2024semantic}. 
For instance, the SWAP planner~\cite{dharmadhikari2023semantics} integrates volumetric exploration with semantics coverage and inspection behavior, while the SB2G~\cite{ginting2024semantic} initially employs geometric coverage behavior and transitions to active semantic search when the robot detects the target object. 
However, these methods are often inefficient in locating the object, particularly in an environment with a large search space, 
because these methods make the robot cover all unexplored areas in the environment until it detects the target object. 
In our work, we use a semantic-guided global planner instead of geometric coverage when the robot does not detect the target object. 
We use relational semantic knowledge to guide the robot in searching promising regions in the environment.


\ph{Semantic knowledge representation}
Semantic knowledge equips robots with richer contextual information. 
Robots gather semantic information from their surroundings using techniques like object detection~\cite{redmon2016you} and semantic segmentation~\cite{he2017mask}. 
This information can be represented in various forms, such as volumetric voxel maps~\cite{grinvald2019volumetric}, neural radiance fields (NeRF)\cite{kundu2022panoptic}, or 3D dynamic scene graphs (DSGs)\cite{rosinol2021kimera}. 
In our work, we use a DSG to model the environment. 
The DSG's structure aligns well with our hierarchical planning method, which utilizes multiple spatial abstraction layers within the DSG.
The DSG can be constructed using floor plans because of the similarity in their structures.  
Moreover, the advancements in LLMs and vision-language models~\cite{chen2022pali, radford2021learning} have led to the new paradigm of scene representations enriched with comprehensive semantic concepts and an extensive open-set object vocabulary~\cite{kerr2023lerf,shafiullah2022clip,jatavallabhula2023conceptfusion}.
Embedding human knowledge about semantic concepts and their interrelations enhances reasoning capabilities in planning.
While open-vocabulary representations can augment our planning approach, we opt for a lightweight prediction model. 
This decision is driven by the need to deploy the model on robots with limited computational power and restricted internet access, especially in industrial inspection sites. 
In this work, we propose an object relationship representation that we call a Relational Semantic Network (RSN). The RSN distills knowledge from LLMs about the likelihood of encountering objects in various regions and refines its predictions based on the robot's observations in the specific environment. 
Given the prior knowledge of the environment, the RSN is trained to predict objects relevant to environment-specific applications and can infer semantically related objects outside of the training data distilled from LLMs.

\textbf{Semantic-guided object-goal navigation} is a prominent research area within embodied AI~\cite{batra2020objectnav}, where advancements are driven by the availability of realistic datasets~\cite{Matterport3D,yadav2023habitat}, simulators~\cite{szot2021habitat,Deitke2020RoboTHORAn}, and competitive challenges~\cite{habitatchallenge2023}. 
Leading approaches train end-to-end reinforcement learning policies (RL)~\cite{maksymets2021thda,wahid2021learning} or combine RL with classical planning methods~\cite{chaplot2020object, ramakrishnan2022poni}. 
Recently, zero-shot approaches have gained popularity through the use of pre-trained language or vision-language models (VLMs), as gathering robotics navigation data is resource-intensive~\cite{majumdar_zson_2023, gadre_cows_2022}. 
While LLMs can be directly queried to make decisions~\cite{singh2023progprompt, xie2023translating}, their planning and reasoning capabilities still raise uncertainties~\cite{valmeekam2022large}. 
Recent state-of-the-art approaches use LLMs and VLMs with a frontier selection approach~\cite{chen2023not,shah2023navigation, zhou2023esc, yokoyama2024vlfm}.
Shah et al. employ LLMs as a guiding heuristic to bias exploration in the direction where the target object may be located~\cite{shah2023navigation}. 
Similarly, Chen et al. present frontier-based exploration with semantic utilities computed using language embeddings~\cite{chen2023not}. 
Yokoyama et al. query VLMs to generate a language-grounded value map by and choose the most promising frontier~\cite{yokoyama2024vlfm}.
More related to our work, Zhou et al. reason over the semantic relationship between objects and rooms and formulate frontier selection with probabilistic soft logic~\cite{zhou2023esc}. 
Our work parallels these methods by incorporating common sense knowledge into the planning, but it is distinct in its approach by focusing on a room-based global planning policy, a slightly different problem setup that is particularly suited for urban inspection scenarios that have access to the spatial configuration of the environment.

\ph{Statement of contribution} 
Our work makes three key contributions in the context of the current literature. 
Unlike state-of-the-art approaches that primarily rely on selecting the best semantic frontier for navigation~\cite{chen2023not,shah2023navigation}, our method introduces a probabilistic planning framework as an MDP, enabling a non-myopic strategy that plans actions over a longer horizon. 
This allows for consideration of future consequences and potential information gains, aiming to compute a globally optimal policy rather than just focusing on immediate short-term outcomes. 
Furthermore, we extend the use of semantic knowledge and the environment information mapped by the robot by introducing a new object-goal navigation problem setup, where the robot leverages the hierarchical structure of environments provided by floor plans. 
Our framework also aims to bridge the gap in applying semantic-based object-goal navigation in practical applications. 
To the best of our knowledge, our work is the first approach that leverages both prior spatial configurations from floor plans and relational semantic knowledge for probabilistic planning in real-world object-goal navigation. 
The practicality of this approach on inspection tasks is demonstrated on hardware, highlighting its relevance and applicability in real-world settings.

\begin{figure*}[h!]
  \centering
  \includegraphics[width=0.95\textwidth]{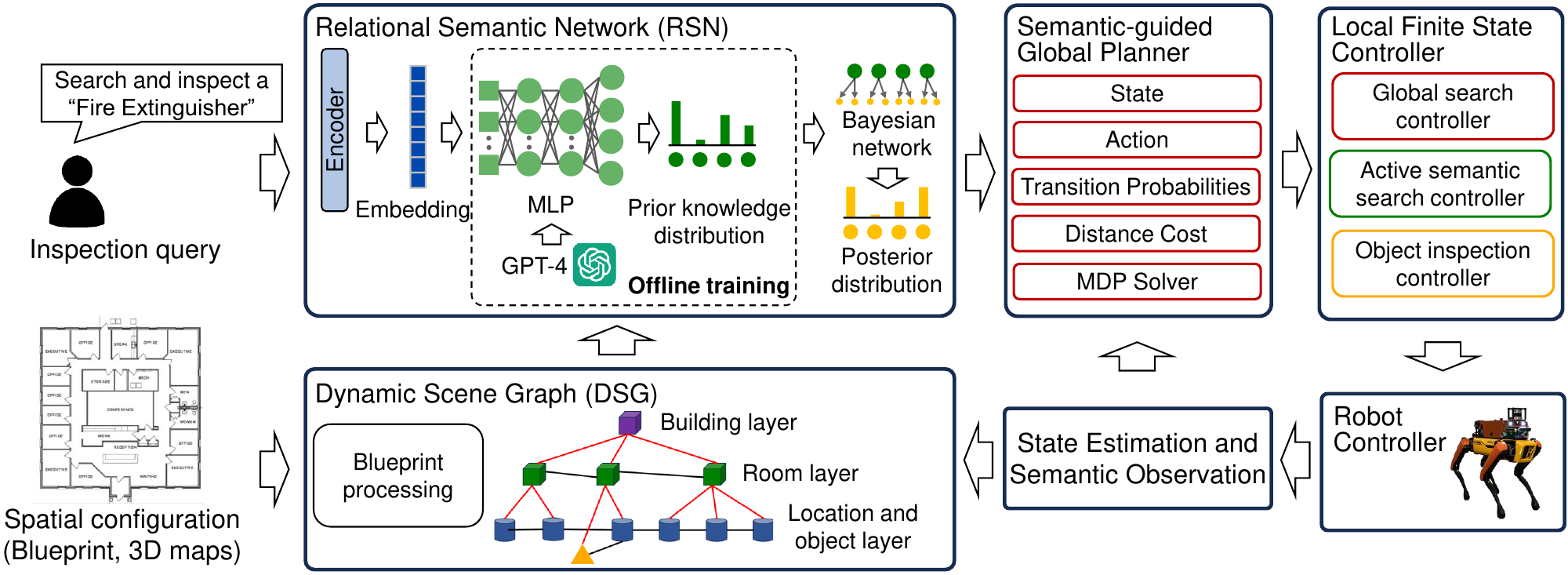}
    \caption{SEEK system architecture. Our method receives an inspection query and builds a Dynamic Scene Graph (DSG) from a given floor plan. The Relational Semantic Network (RSN) predicts the probability of finding the target object in every room. 
    The semantic-guided global planner uses this prediction to compute an optimal global search policy. Finally, the local finite state controller executes the global policy and switches to an active semantic search controller when the robot sees the object.}
  \label{fig:sysarch}
\end{figure*}

\section{Object-Goal Navigation with Relational Semantic Knowledge}
\subsection{Problem Statement}
We first define the problem of object-goal navigation. Let $x_k \in X$, $u_k \in U$, and $z_k \in Z$ denote the robot state, action, and observation at time step $k$, respectively. 
The process model $x_{k+1} = f(x_k, u_k)$ encodes the system dynamics. 
The robot generates action $u_k$ according to a policy $\pi$. 
The robot observes and maps various objects in the environment as $y=\{y_{1}, \ldots, y_{N_k}\}$ with $y_{i} \in Y$, where $N_k$ represents the number of observed objects until time step $k$.
Each object state $y_{i} \coloneqq (y_{i}^p, y_{i}^q, y_{i}^l)$ 
captures the object's position $y_{i}^p$, 
orientation $y_{i}^q$, and semantic class $y_{i}^l \in L$ of the object.

In partially observable environments, the true state of $y$ is unknown to the robot. 
If an object $y_{i}$ is visible from the current robot state $x_k$, the robot can observe the object. 
The observation $z_k \coloneqq (z_k^p, z_k^q, z_k^l, z_k^s)$ consists of the object's measured position $z_k^p$ and orientation $z_k^q$, detected class $z_k^l$, and detection confidence score $z_k^s$.
The observation model $z_k = h(x_k, y_k)$ encodes the relationship between $(x_k, y_k)$ and $z_k$. 
The estimated state of the objects $y$ and the robot state is represented as a belief $b_k \coloneqq p(x_k, y \mid \mathcal{H}_k)$, where $\mathcal{H}_k$ denotes the history of past observations and actions. 
We use $b_k$ as the basis for decision $u_k = \pi(b_k)$. 
The belief evolution model $b_{k+1} = \tau(b_k, u_k, z_{k+1})$ updates $b_k$ and can be computed recursively~\cite{thrun2005probabilistic}.

\ph{Object-goal navigation} 
The robot's objective is to search and navigate to a target object $y_G \in Y$ for object inspection. 
The target object is described as the object semantic class $y^l$. 
The robot successfully navigates to $y_G$ when the robot position $x_k^p$ is close to the target object $\|x_k^p - y_G^p\| < \epsilon$ and triggers a finish signal $u^{f}$. 
The robot must efficiently search for the object by minimizing the distance traveled to the target object. 

\ph{Problem 1 (Object-Goal Navigation)}
Given a target object $y_G$, initial robot state $x_0$, find an optimal object search policy $\pi^*$:
\begin{align}
    \label{eq:problem_statement}
    \pi^* = &\argmin_{\pi \in \Pi} \sum_{k=0}^{T-1} \|x_{k+1}-x_{k}\|\\ \nonumber
    \text{s.t.~} & \|x_T^p - y_G^p\| < \epsilon, u_T = u^f, \\ \nonumber
    & x_{k+1} = f(x_k, \pi(b_k)),\\ \nonumber
    & z_k \sim p(z_k \mid x_k, y_k) ,\\ \nonumber
    & b_{k+1}=\tau(b_k, \pi(b_k), z_{k}).
\end{align}

\subsection{Semantic Knowledge Representation} 
We use two sources of prior semantic knowledge to guide the robot in locating the target object. 
The first source of prior knowledge comes from the spatial configuration of the environment.
This information can come from floor plans and  Building Information Models (BIM) commonly available in industrial inspection tasks. 
The spatial configuration provides the names of regions or rooms in the environment that are semantically meaningful in locating the target object. 
The second source of prior knowledge is the semantic relationship between target objects and room names. 
The semantic relationship informs which rooms the target object is usually located in.
We maintain spatial configuration in a Dynamic Scene Graph (DSG) and semantic knowledge relationship in a Relational Semantic Network (RSN).

\textbf{Dynamic Scene Graph (DSG)}  $\mathcal{G} = (\mathcal{V},\mathcal{E})$ is a hierarchical representation of the environment. 
The DSG nodes $\mathcal{V}$ can be partitioned into $N$ layers with $\mathcal{V} = \cup_{i=1}^N \mathcal{V}_i$. 
Each node $v \in \mathcal{V}_i$ at layer $i$ can only share an edge with at most one parent node in the layer above $\mathcal{V}_{i+1}$ and can only share edges with nodes in the same or adjacent layers $\mathcal{V}_{i-1}, \mathcal{V}_{i}, \mathcal{V}_{i+1}$. 
The node $v \in \mathcal{V}$ encodes the node position $v^p \in \mathbb{R}^3$, orientation $v^q \in SO(3)$, and the semantic class $v^l \in L$. 
The edges on the same layers represent direct spatial connectivity between the nodes, and the edges between different layers represent how the nodes are spatially grouped together.
In the urban environment, the hierarchy levels are named location/objects, rooms or regions, and buildings in ascending order. 
Prior to the robot deployment, the DSG nodes on the room-level $\mathcal{V}_2$ and above are initialized from the prior knowledge of the spatial configuration.

\textbf{Relational Semantic Network (RSN)} $\mathcal{F}$ is a Bayesian network that estimates the probability of finding target objects in the environment $P(y_G) = \mathcal{F}(y_G, \mathcal{G})$. 
The network estimates the probability of finding $y_G$ in every room $P(y_G) \in \mathbb{R}^{|\mathcal{V}_2|}$. 
The RSN is initialized using relational semantic knowledge on the object occurrences in different semantic types of rooms. 
The probability $P(y_G)$ is updated as the robot discovers the target object in the environment across planning episodes $P(y_G)_{k+1} = P(y_G \mid z_{k+1})_k$, where $P(\cdot)_k$ represents object probability at time step $k$.

\subsection{Planning with Relational Semantic Knowledge}
We propose a probabilistic and hierarchical planning approach for the object-goal navigation (\autoref{fig:sysarch}). 
First, the global planner computes a policy $\pi^g$ for the robot to search for the target object across all rooms $\mathcal{V}_2$ using relational semantic knowledge encoded in an RSN. 
Then, the local planner generates a robot control policy $\pi$ that moves the robot to a room $v_i=\pi^g(x_k)$ while searching for $y_G$ in a local area around the robot. 
This approach enables a more efficient global search by directing the robot to search more promising rooms informed by relational semantic knowledge.

\textbf{Global planner} solves the problem of navigating through a set of rooms $\mathcal{V}_2$ with the objective of locating the target object $y_G$. 
We formulate this problem as a  Markov Decision Process (MDP).
An MDP is defined by the tuple $(\mathbb{X}, \mathbb{U}, P, C)$, which represents the state space, action space, transition probabilities, and cost function, respectively. 
The state $\mathcal{X} \in \mathbb{X}$ comprises the current room where the robot is within $\mathcal{V}_2$, and a goal state $\mathcal{X}^G$, where the robot finds $y_G$. 
The action space $\mathbb{U}$ is the set of rooms $\mathcal{V}_2$ that the robot can go to $\mathbb{U}^{\text{visit}}$ or search locally $\mathbb{U}^{\text{search}}$. 
The size of the action space is $|\mathbb{U}| = 2|\mathcal{V}_2|$. 
The action space $\mathbb{U}$ consists of two types of actions: moving to a room $\mathbb{U}^{\text{move}}$ and searching within the current room $\mathbb{U}^{\text{search}}$. 
The size of the action space $\mathbb{U}$ is $2|\mathcal{V}_2|$, comprising one action to move to every room and one search action within each room.
The transition probabilities $P$ consist transition probabilities between rooms $P(\mathcal{X}_{k+1}=v_j \mid \mathcal{X}_{k}=v_i, \mathcal{U}_k)$ and transition probabilities to the goal state $P(\mathcal{X}_{k+1}=\mathcal{X}^G \mid \mathcal{X}_{k}=v_i, \mathcal{U}_k)$. 
The transition probabilities to the goal state are estimated by the RSN $P(y_G)$. 
The cost function $C$ in our formulation is the expected traversal distance between rooms or the expected distance to search the room of the current robot location. 
Finally, to compute for an optimal policy for $\pi^g$, we minimize the expected cost-to-go $J(\mathcal{X})$ such that
\begin{align}
    \label{eq:mdp}
    J(\mathcal{X}_i) &= \min_{\pi \in \Pi^g} C(\mathcal{X}_i, \mathcal{U}) + J(\mathcal{X}_{j})  \\ \nonumber
    \pi^g &= \argmin_{\pi \in \Pi^g} J(\mathcal{X}_i),
\end{align}
where $\mathcal{X}_i$ and $\mathcal{X}_j$ denote the current state and the next state, respectively.

\textbf{Local planner} $\pi^l$ generates a robot control policy $\pi$ based on the global policy $\pi^g$ and the current belief $b_k$. 
It switches between three controllers to search for and navigate to the target object. 
The first controller $\rho_{\text{nav}}$ generates a navigation policy to visit a new room or search the current room based on the action prescribed by $\pi^g$. 
When the robot detects a possible sign of the target object using an object detector~\cite{redmon2016you}, the local planner switches to the active semantic search controller $\rho_{\text{search}}$. 
This controller is triggered when the robot belief $b_k$ of an object is within the belief set $B^{\text{search}}$. 
The controller actively gathers new observations to increase its confidence and drive $b_k$ to reduce the target object's pose and semantic detection uncertainty. 
If the belief of target object $b_k$ has low uncertainty, $b_k \in B^{\text{inspect}}$, the robot uses an inspection controller $\rho_{\text{inspect}}$ to navigate the robot to the target object within a radius $\epsilon$ and trigger the finish signal $u^f$.
In summary, the local planner serves as a finite state controller to generate robot policy $\pi_k = \pi^l(\pi^g(x_k), b_k)$.

\section{Semantic-guided Object-Goal Navigation}
Having outlined our approach for solving object-goal navigation with relational semantic knowledge, we discuss how we build the DSG and RSN and compute the object-goal navigation policy. 
Our approach consists of an offline initialization of the DSG and RSN and an online object-goal planning. 
Before deploying the robot for the task, we generate the DSG from floor plans (\autoref{subsec:dsg}). 
We then train the RSN to predict object occurrences to guide the robot in searching for objects that can be found in the environment (\autoref{subsec:rsn}). 
During the deployment, when the robot receives a request to inspect a target object $y_G$, the global planner computes the global policy that minimizes the expected travel distance to search for $y_G$ through the rooms or regions in the environment (\autoref{subsec:gp}). 
Finally, the local finite state controller selects a control policy for the robot based on the estimated belief of the target object (\autoref{subsec:local_controller}). 
As the robot observes the environment, it recomputes its global and local policy and saves the DSG and RSN for future deployments. 
\autoref{fig:sysarch} summarizes the architecture of SEEK.

\subsection{Dynamic Scene Graph (DSG) Generation}
\label{subsec:dsg}
We generate the DSG of the environment using the environment's floor plan. Floor plans provide geometric and semantic information about the room/region nodes $\mathcal{V}_2$. 
For the location nodes $\mathcal{V}_1$, we sample the nodes uniformly across all the open space based on the floor plan to ensure full room coverage. 
We compute the Euclidean distance between $\mathcal{V}_1$ and store the information on the edges. 

When the robot is being deployed in the environment, the robot updates the DSG. 
First, the robot self-localizes using its LIDAR-based map to the floor plan~\cite{reinke2022locus}. 
The robot navigates in the environment and updates the traversability risks~\cite{fan2021step} between the location nodes. 
The connectivity can change across deployments due to the evolving object configuration in the environment over time, which is not described in the floor plans. 
During the deployment and in our experiments, we do not consider changes in the connectivity between rooms and buildings. 
When the robot observes and localizes an object, 
a new object node is created on the first layer and connected to the nearest location node. 
The updated DSG is used for path planning and stored for future deployments.

\subsection{Relational Semantic Network (RSN)}
\label{subsec:rsn}
We present a compact semantic information representation for object-goal navigation. 
RSN is trained using common-sense knowledge to estimate the probability of the object's presence in various room types. 
The RSN is specifically designed to extract semantic relationship knowledge from LLMs, as in industrial inspection scenarios, the robot typically lacks remote access to LLMs and has limited computational resources to host large models onboard. 
We present how we build, train, and update the RSN.

The RSN consists of a sequence of three networks that receive the target object semantic class $y_G^l$ in text and output the probability of finding $y_G$ across rooms $P(y_G)=\mathcal{F}(y_G, \mathcal{G})$. 
The first network is a pre-trained text encoder that maps the target object class to a text embedding. 
A text embedding model or vision language embedding model that captures semantic similarity between objects can be used as the text encoder. 
We use a small BGE embedding model~\cite{bge_embedding}, a small-scale model with competitive performances in massive text embedding benchmark~\cite{muennighoff2022mteb}. 
The text embedding is then passed through a multi-layer perceptron (MLP) that estimates the probability of finding the target object in every room type. 
The final network is a Bayesian network that estimates the probability of finding the target object $P(y_G)$ in every room $\mathcal{V}_2$ given the MLP estimates and past observations.

The MLP has three hidden layers and outputs two types of information: the probability of finding an object on 27 different room types and the probability of finding an object without searching the room carefully $\mathbb{U}^{\text{search}}$. 
The room types consist of a common semantic class of rooms in indoor environments and are extracted from the Matterport3D dataset. 
The probability of finding an object without a local room search estimates how easy it is to find $y_G$ when entering a room for the first time. 
More prominent and highly visible objects, such as fridges and fire extinguishers, have a higher value compared to smaller objects that are possibly placed in cluttered spaces, such as coffee mugs or laptops. 

We train the MLP using 100 object names that are usually queried for object-goal navigation in home and urban buildings. 
We train an MLP to enable RSN to predict semantic relationships for semantically similar object names outside the training data. 
To generate the training data, we first query GPT-4 to get a list of the objects. 
Then, for the given object list, we again query GPT-4 to estimate the probability of finding the object in every room type and the probability of finding an object without a careful room search. 
We query the model for every five objects due to the current limits of the output token. 
Using the dataset created by GPT-4 and text embedding of the object name, we train the MLP. 
Using text embedding helps MLP to estimate the probability of objects outside of the training data because similar objects are located close to each other in the embedding space. 
We evaluate the prediction performance of the MLP outside of the training dataset with different text embedding models in \hyperref[app:test_embedding]{Appendix C}.

The Bayesian network updates $P(y_G)$ when there are new environmental observations. 
Updating $P(y_G)$ moves the probability closer to the true distribution of the object placements in the environment. 
While the MLP trained using internet data can help to initially estimate $P(y_G)$, updating $P(y_G)$ is crucial because the environment has variations in object placement. 
We use a naive Bayes model as we can assume conditional independence of the object observation on every room in the environment \cite{kochenderfer2022algorithms}.
The root node of the model is the prior distribution of $y_G$ from the previous inference step. The evidence node of the model is a binary value of whether the robot sees the target object in every room after a local search. This conditional distribution is estimated from the MLP. 
We perform exact inference of $P(y_G)$ using the rule of conditional probability and marginalization.

\ph{Extension to open vocabulary prediction} 
The RSN is designed to encode semantic relationships in a model that can be deployed on the robot with limited computational resources and no remote access to LLMs. 
The network is trained to predict $P(y_G)$ of objects that are expected to be queried for environment-specific applications and semantically similar objects from the training dataset. 
Given access to an LLM on the robot or remotely, the RSN can be adapted to an open vocabulary setting by replacing the text encoder and MLP network with a direct query to LLMs.

\begin{figure}[t]
    \centering
    \includegraphics[width=0.3\textwidth]{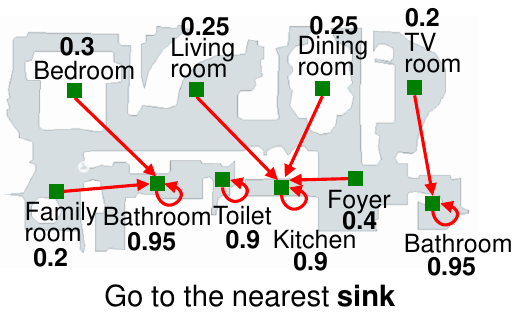}
    \caption{A visualization of the global search policy. The RSN predicts the probability of finding a sink in every room. The policy computes the best action from each room to locate the nearest sink. The circular arrow represents the local search. }
    \label{fig:policy_example}
\end{figure}
\vspace{-5pt}

\subsection{Semantic-Guided Global Planner} 
\label{subsec:gp}
Given the DSG and the RSN, we have all the information to solve the semantic-guided global planning problem through the MDP formulation. 
The state space $\mathbb{X}$ and action space $\mathbb{U}$ are extracted from the set of room nodes $\mathcal{V}_2$ in the DSG. 
To specify the transition probabilities $P$, we first set the transition probabilities to reach the goal state $P(\mathcal{X}_{k+1}=\mathcal{X}^G \mid \mathcal{X}_k, \mathcal{U}_k)$ by querying the RSN. The transition between rooms is deterministic since the DSG $\mathcal{G}$ is a connected graph. 
Then we can set the transition probabilities to reach a new room after taking $\mathbb{U}^{move}$ action and staying in the same room after taking $\mathbb{U}^{\text{search}}$ action as $1 - P(\mathcal{X}_{k+1}=\mathcal{X}^G \mid \mathcal{X}_k, \mathcal{U}_k)$. 
To specify the cost function $C$, we compute the shortest travel distance to visit rooms using A$^*$ search~\cite{duchovn2014path} on the DSG 
and compute the total distance to visit location nodes in every room. 
Given that the room connectivity is assumed not to change during the run, the cost function $C$ can be computed offline and stored as a look-up table for planning. 
The cost function can be recomputed during the run if we consider a robot deployment with changing room connectivity. The cost function can be recomputed between the global planning computation.

We solve the MDP planning problem in \autoref{eq:mdp} in real-time on the robot. 
Given a relatively small state and action space of our problem ($|\mathbb{S}| < 100$ and $|\mathbb{U}| < 200$), we use value iteration to compute the optimal policy $\pi^G$~\cite{kochenderfer2022algorithms}. 
When querying a global action $\mathcal{U}$ from $\pi^g(x_k)$, the robot chooses the action associated with the state of the closest room node from the robot pose. 
\autoref{fig:policy_example} illustrates an example of a global policy to navigate to a target object.

We update $\pi^G$ in a receding horizon manner. After executing a global action $\mathcal{U}_i$, the robot will have an updated belief of the target object distribution given a new observation $p(y_G \mid z_{k+1})$. 
We update the transition probabilities $P$ and solve for a new $\pi^G$.

\subsection{Local Finite State Controller}
\label{subsec:local_controller}
We use three different controllers that execute the global policy $\pi^g$, search, and navigate to the object. We use the Semantic Belief Behavior Graph (SB2G) framework to implement and define the controller switching~\cite{ginting2024semantic}. 

\textbf{Global search controller} $\rho_{\text{nav}}$ computes a robot policy according to the global policy $\pi^g(x_k)$. If the global action is to move to a new room $\mathbb{U}^{\text{move}}$, the controller computes the shortest path to the room on the graph and passes the path to a risk-aware MPC controller that computes a robot trajectory~\cite{fan2021step}. 
If the global action is to search the current room, the controller plans a coverage path that explores the obstacle-free area in the room~\cite{bouman2022adaptive} and passes the path to the MPC controller. 

\textbf{Active semantic search controller} $\rho_{\text{search}}$ plans a path that maximizes the information gain of observing a potential target object. 
We follow the active semantic search algorithm presented in the SB2G~\cite{ginting2024semantic}.

\textbf{Object inspection controller} $\rho_{\text{inspect}}$ computes a policy to move the robot as close as possible to the detected target object. The controller chooses a target robot pose on an obstacle-free area closest to the target object. 
The target pose is then given to the MPC controller.

\section{Experimental Results} 
We evaluate the performance of our approach in representative simulation and on hardware. 
We first evaluate the performance of semantic-guided global planner against other methods in the Habitat simulator~\cite{savva2019habitat}. 
Then, we analyze the performance of SEEK in a representative office environment in the ROS Gazebo simulator. 
Finally, we deploy SEEK on a legged robot performing real-world inspection scenarios in an office building.

\subsection{Global Planner Evaluation}

We demonstrate the performance of our approach in the Habitat simulator using the Matterport3D dataset~\cite{Matterport3D}. 
We are not training our RSN with any data from the dataset. 
The Matterport3D dataset provides a hierarchical scene configuration that we use to build the DSG. 
The robot is tasked to navigate to the closest object instance~\cite{batra2020objectnav}.
We sample five scenes and ten different objects from the Matterport3D dataset. 
The simulations were performed on a laptop with an Intel i9-11950H CPU.

We evaluate SEEK against four approaches: 
\begin{enumerate}
    \item \textbf{Semantic Utility} method selects a room that has the highest semantic utility.
    We adopt the idea of selecting frontier with semantic utility~\cite{chen2023not} to the room selection problem. 
    The robot chooses an unvisited room with the highest utility $\mathbb{U}_{i,sem} = 1  / (\mathrm{dist_{sem}}(y_G, v_i)~\cdot~\mathrm{dist}(x_k, v_i))$. The semantic distance $\mathrm{dist_{sem}}(y_G, v_i)$ is the distance between the target object and the room name in the word embedding space. The intuition behind using semantic distance is that objects and room names that are more semantically related have closer distances in the embedding space, reflecting the likelihood that the object is commonly found in that room. We use BERT embedding, the best-reported model for semantic frontier method~\cite{chen2023not}. 
    The geodetic distance $\mathrm{dist}(x_k, v_i)$ is the shortest path length from the current robot position to the centroid of the room.
    \item \textbf{GPT-4 Planner} provides the sequence of action for the robot based on the given context in the query. We provide the exact information as we give to our RSN and global planner in text, including the list of rooms, path lengths between rooms, robot state, action space, and inspection objective. We use the Chain-of-Thought prompting technique~\cite{wei2022chain} to guide the robot to reason where to find the object. The prompt that we use is provided in Appendix \ref{app:prompt}.
    \item \textbf{Room Coverage} selects the next room greedily based on the path distance between the current robot pose to unvisited rooms. This method serves as a baseline of an object search method without semantic guidance.
    \item \textbf{Random} policy chooses a sequence of rooms to be searched at random. The random baseline is used to evaluate the benefit of other approaches.
\end{enumerate}
For every method, we run 50 simulations in the Matterport3D dataset. 
To assess the benefit of relational semantic knowledge, in this experiment, our method does not remember the target object location and any information from previous runs.

\begin{table}[t!]
\renewcommand{\arraystretch}{1.27}
\caption{ \textbf{Global planning  results}. We report the mean and standard deviation of the Success weighted by inverse path length (SPL). Results are separated into two categories: \textit{Fixed Objects} and \textit{Movable Objects}, to demonstrate the effectiveness of different methods on different types of objects.}

\centering
\begin{tabular}{@{}lcc@{}}
\toprule
\textbf{Methods} & \multicolumn{2}{c}{\textbf{SPL Mean (Std. Deviation)}} \\
& \textbf{Fixed Objects} & \textbf{Movable Objects} \\
\midrule
\textbf{SEEK (Ours)} & \textbf{0.96 (0.12)} & \textbf{0.84 (0.23)} \\
Semantic Utility & 0.74 (0.27) & 0.70 (0.27) \\
GPT-4 Planner & 0.84 (0.26) & 0.81 (0.23) \\
\textit{Room Coverage} & 0.77 (0.25) & 0.65 (0.27) \\
\textit{Random} & 0.65 (0.35) & 0.68 (0.32) \\
\bottomrule
\end{tabular}
\label{tab:spl_habitat_results}
\vspace{-4pt}
\end{table}

We evaluate the performance of our approach with the SPL (Success weighted by Path Length) metric, a standard metric to measure object-goal navigation performance~\cite{batra2020objectnav}. 
The SPL is defined as
\begin{align}
    SPL = \frac{1}{N_{run}} \sum_{i=1}^{N_{run}} S_{i} \frac{l}{\max(p_{i}, l)}.
\end{align}
In this definition, \( S_i \in \{0,1\} \) denotes whether the \(i\)-th run was successful (1) or not (0), \( l_i \) represents the length of the shortest possible path to the nearest instance of the target object, and \( p_i \) indicates the length of the path actually taken by the agent during the \(i\)-th run. 
In our experiments, given the target object's presence in the environment and unlimited simulation time, all methods are capable of locating the target object. This is because the action space of each method enables the robot to visit and search every room. Therefore, in these conditions, $S_i$ is always equal to 1.

\autoref{tab:spl_habitat_results} presents the performance of various methods at locating fixed objects (e.g., bed, sink) and movable objects (e.g., clothes, towel).
Our method outperforms other methods on both object categories. 
Our approach is significantly better at locating fixed objects, which can be attributed to their distinct association with specific room types. 
In contrast, searching for movable objects is relatively more challenging as they can be found in various rooms.

We observed that the Semantic Utility method has lower efficiency than our method. The semantic utility cost on room selection places greater emphasis on geodetic distance rather than interclass distance. 
Adjusting the weight balance between these factors based on the size of the environment could potentially enhance performance, although such tuning may not always be desirable.

GPT-4 Planner ranks second in simulation performance. 
It demonstrates competent reasoning in object search tasks. 
However, its limitations become apparent in probabilistic and spatial reasoning. 
GPT-4 Planner's rule-based approach struggles with probabilistic reasoning, which is crucial in scenarios with uncertainties about the environment and object locations. 
For example, when the robot is tasked with finding clothes, the planner moves the robot to the nearest bathroom instead of prioritizing the bedroom and bathroom, which are slightly farther away but jointly have a higher probability of containing clothes. 
This indicates a lack of inherent understanding of spatial relationships in GPT-4 Planner's reasoning.

\begin{figure}[t]
    \centering
    \includegraphics[width=0.48\textwidth]{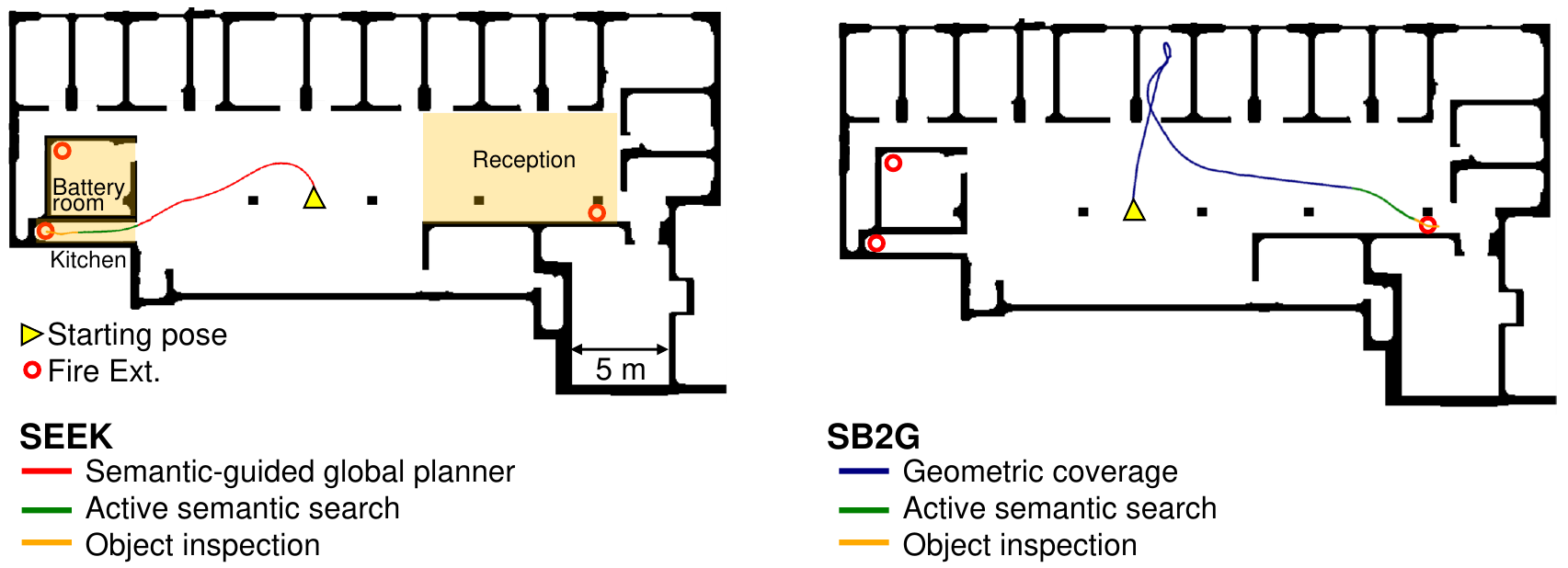}
    \caption{Robot paths comparison of object-goal navigation with SEEK and SB2G in the Gazebo simulator. Our global planner guides the robot to the kitchen, where a fire extinguisher is usually located. 
    }
    \label{fig:sim_path}
\end{figure}
\begin{figure}[t]
    \centering
    \includegraphics[width=0.48\textwidth]{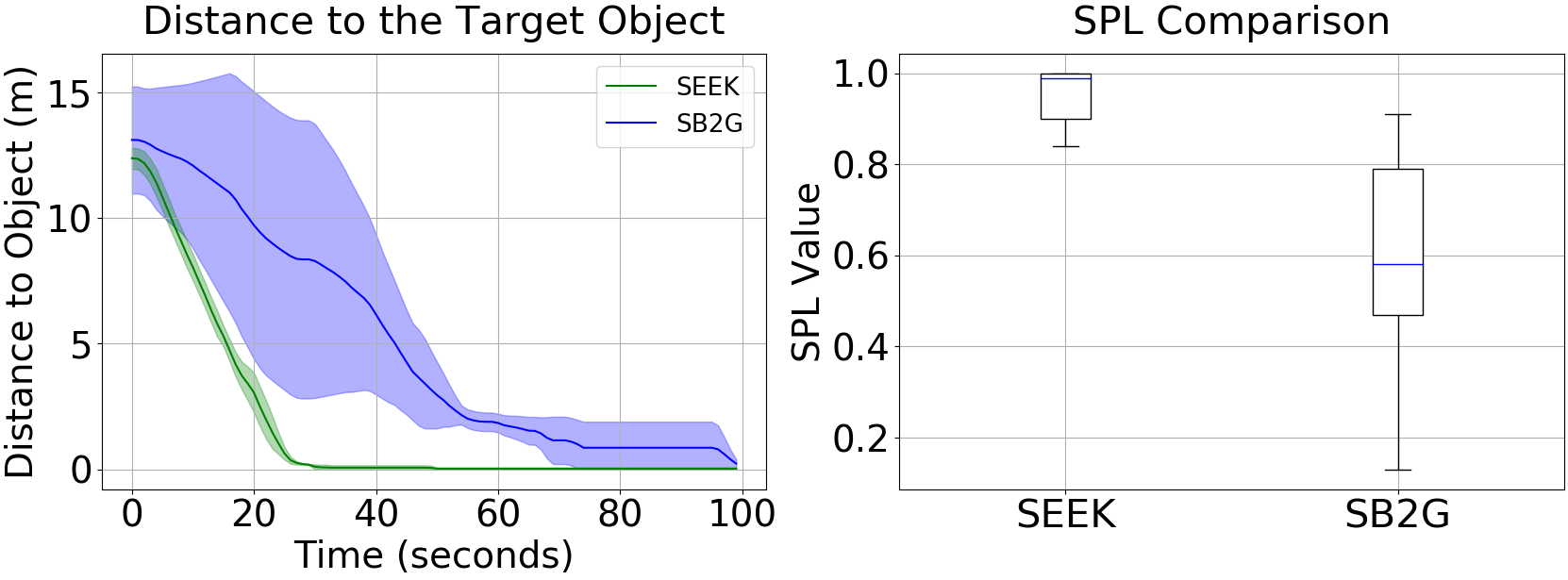}
    \caption{The left plot compares how fast the robot searches and navigates to the target object from the same starting location using SEEK and SB2G. 
    The shades on the line plot represent the standard deviation across three runs with the same starting locations. 
    The right box plot reports the SPL metric across seven runs with different starting location.
    }
    \label{fig:sim_quant}
\end{figure}

\begin{figure}[t]
    \centering
    \includegraphics[width=0.48\textwidth]{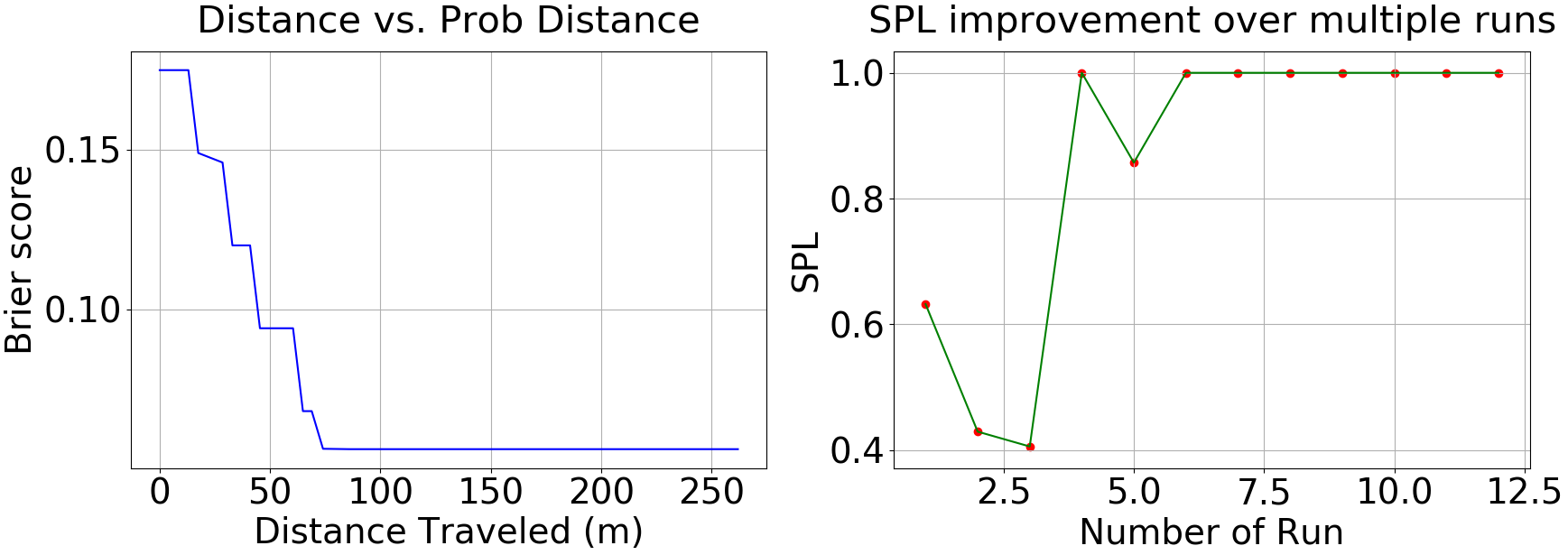}
    \caption{The improvement of the RSN prediction accuracy (left) and the SPL (right) as the robot observes different target objects across the runs. A lower Brier score indicates better prediction accuracy. 
    }
    \label{fig:sim_improve}
\end{figure}

\begin{figure*}[h!]
  \centering
\includegraphics[width=1.0\textwidth]{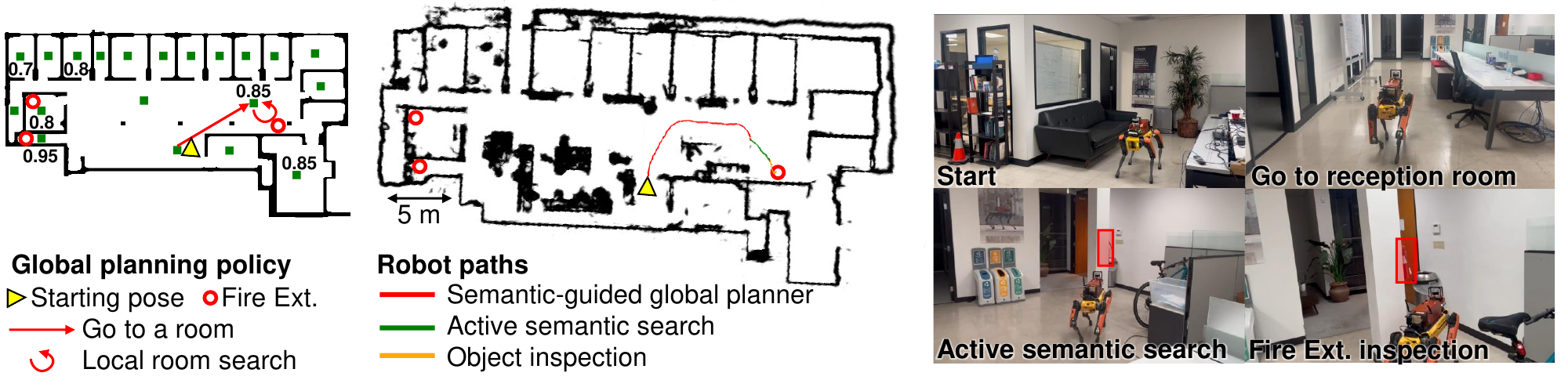}
    \caption{\textbf{Experiment results on hardware.} The left figure visualizes the global planning policy executed by the robot. We mark the rooms with a high probability of finding the target object based on the RSN prediction. The middle figure plots the path taken by the robot. The right figure shows a third-person view of the robot on different phases of the inspection.}
  \label{fig:hw_results}
\end{figure*}

\subsection{Simulation Study}
We evaluate SEEK on object-goal navigation tasks in a simulated environment of an office space.
We use the ROS Gazebo simulator with the Spot model from Boston Dynamics. 
In this experiment, the robot is tasked to search and navigate to the nearest fire extinguisher on the first set of simulations and to the coffee mug on the second simulation. 
The environment consists of $21$ rooms based on the office floor plan (\autoref{fig:sim_path}).
The placement of fire extinguishers and coffee mugs is based on the real placement in our office. 
We simulate the object detection and localization using a semantic observation model~\cite{atanasov2016localization}, with model parameters based on empirical data from our hardware experiments in the office. 
The RSN network is not trained in any of the object placement distributions. 
The entire software stack and simulations were performed on a laptop with an Intel i9 CPU.

We first evaluate the performance of SEEK to find the nearest fire extinguisher across $7$ runs with different starting locations. 
We compare our approach against the SB2G method that uses geometric coverage to explore the environment until it detects the target object. 
The SPL performance comparison and a sample of the robot path are presented in~\autoref{fig:sim_quant} and \autoref{fig:sim_path}. 
Our approach reasons what type of rooms in the office that usually have a fire extinguisher. 
The semantic-guided global planner successfully directs the robot to the nearest room with a fire extinguisher. 
Compared to the SB2G method, SEEK significantly reduces the distance traveled by the robot to find the object.  
On multiple runs starting from the same location, the distance variation is much lower compared to SB2G, which uses geometric coverage that can stochastically produce different paths to maximize space coverage.
On various starting locations, our method  has a higher SPL than SB2G, validating the benefit of the semantic-guided global planner.

We then demonstrate how SEEK's planning policy improves over multiple runs to locate a coffee mug in the office. 
We perform 12 runs on three different starting locations. 
In \autoref{fig:sim_improve}, we present the SPL performance and the Brier score~\cite{blattenberger1985separating} of the predicted target object location by the RSN with the true object location in the environment. 
The object search performance improves over the runs as the robot observes different coffee mugs in the environment. 
The accuracy of the target object probability estimate is constantly increasing, highlighted by the decreasing Brier score over time. 
The RSN updates the target object probability using the observation on previous runs.

We evaluate the performance of our approach when the target object is placed in unexpected locations. We present the results and discussion in \hyperref[app:cornercases]{Appendix B}.

\subsection{Hardware Results}
We validate our approach by deploying SEEK on the Boston Dynamics Spot legged robot for an inspection task within an office building.
The robot is equipped with a LIDAR and three cameras for navigation and semantic observation. 
The robot uses a LIDAR-based SLAM method for localization~\cite{reinke2022locus} and a YOLO-based model~\cite{redmon2016you} for object detection. 
We build the DSG using the office floor plan and use the RSN without any fine-tuning to the office data. 
The object detection model runs on an NVIDIA Jetson AGX Xavier computer, and the rest of the stack runs on an Intel i7 computer on the robot.
The robot receives an object-goal inspection query through a remote user interface. 
In this experiment, the robot is tasked to locate and go to a fire extinguisher, a commonly inspected object, due to its critical role in safety and emergency response. 
There are three fire extinguishers located in the office, including in the kitchen area, battery charging room, and the office receptionist.

Using SEEK, the robot efficiently searches and navigates to the nearest fire extinguisher (\autoref{fig:hw_results}). 
The global policy directs the robot to search rooms/regions where fire extinguishers are usually located. 
In the experiment presented in ~\autoref{fig:hw_results}, our approach is able to navigate to the closest fire extinguisher with $SPL=0.93$. 
SEEK reduces the inspection time by directing the robot to search in rooms that have a higher probability of finding fire extinguishers. 
By distilling the semantic knowledge to the RSN, the robot does not need to access semantic knowledge directly from LLMs or large foundational models that require significant computational resources onboard or access to the internet. 
The RSN inference and the global planning can be performed in real-time, making our approach feasible for real-world object inspection.

\section{Conclusion} 
\label{sec:conclusion}
We have presented SEEK, a framework for object-goal navigation in real-world inspection tasks.
SEEK uses prior spatial configuration of the environment and relational semantic knowledge to guide the robot to search target objects in the environment. 
Using relational semantic knowledge, we presented a novel probabilistic planning algorithm for object-goal navigation.
Through simulations and hardware experiments, we demonstrated the efficacy and practicality of our approach for real-world object inspection.

For future work, we plan to use the knowledge of semantic relationships between objects to augment the RSN's prediction capability. 
Images and detected objects in different parts of the environment can be encoded using vision-language models to provide richer contexts to the RSN. 
For industrial applications, the hierarchical information of objects contained in the BIM model can be used to train the semantic relationship~\cite{kayhani2023semantic}. 
This extension could improve the local object search and support open-vocabulary object-goal navigation in real-world inspection tasks.

\section*{Acknowledgments}
We gratefully acknowledge Shayegan Omidshafiei for his invaluable discussions and insights, which greatly contributed to the refinement of this work.

\bibliographystyle{plainnat}
\bibliography{references}

\clearpage 
\appendix

\section*{A. GPT-4 Planner Prompt}
\label{app:prompt}
The prompt template used in the simulation experiment.
\begin{mdframed}
\textbf{Context}\\
We want to guide a robot to search and inspect a target object in a house.
We have some prior information from the floor plan of the list of room names and the centroid position of the room.
We also have a distance matrix of the robot path length to move between the rooms. 
We also have the history of action, what room the robot has visited or searched carefully, and the history of the observation, what the robot sees after searching carefully.
The action that we can give to the robot is:
\begin{enumerate}
    \item Go to a room
    \item Search a room locally
\end{enumerate}
If the robot sees the target object, the robot can autonomously navigate to and inspect the target object. \\
Can you guide the robot to search and navigate to the target object efficiently by giving a robot a policy?

\textbf{Input} \\
Following is the information that we have
\begin{enumerate}
    \item Target object:
    \item List of rooms:
Region id:x, category:x, center:[x   y z ]
    \item Distance matrix following the id of the room
    \item Current robot location:
    \item Past action: 
    \item Past observation:
\end{enumerate}
\textbf{Output}\\
Please explain your reason to guide the robot that way and finally provide the sequence action for the robot until it finds the target object
\end{mdframed}

\section*{B. Planning Evaluation for Target Objects Located in Unexpected Locations}
\label{app:cornercases}
\begin{figure}[t]
    \centering
    \includegraphics[width=0.48\textwidth]{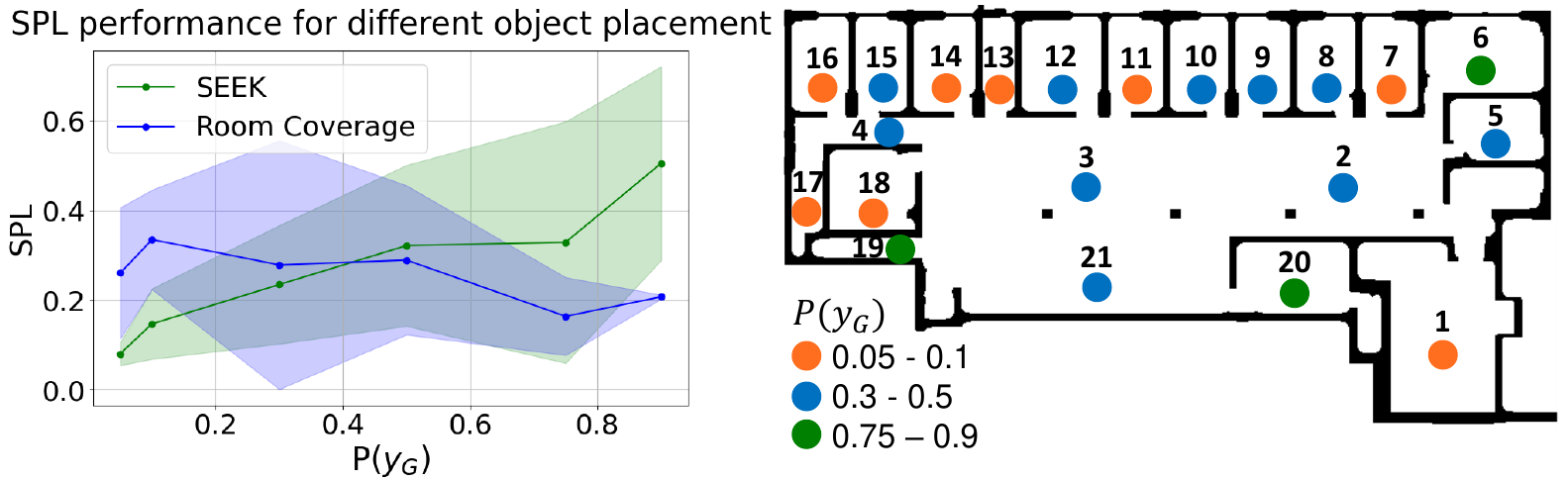}
    \caption{The left plot shows the decrease of the SPL performance as the target object, a coffee mug, is placed in unexpected rooms. The performance is compared with the \textit{Room Coverage} method, and the shades represent the standard deviation. 
    The right map represents the office map labeled with the room number and the predicted probability of finding a coffee mug in every room.
    }
    \label{fig:different_placements}
    \vspace{-7pt}
\end{figure}

We evaluate the performance of SEEK when the target object is located in unexpected locations. 
In this simulation, the robot is tasked to find a coffee mug in an office environment (\autoref{fig:different_placements}). 
To make the object search more challenging, only one coffee is mug placed in the environment. 
We evaluate the runs on 18 different placements with three different starting locations. 
The placement of the coffee mug varies from the most expected location (e.g., kitchen) to the least expected location (e.g., PPE storage room). 
We also simulate the \textit{Room Coverage} method as a benchmark for object search without semantic prior knowledge. 
The office map in \autoref{fig:different_placements} summarizes the predicted probability $P(y_G)$ of the object's presence in every room.

\begin{figure}[t]
    \centering
    \includegraphics[width=0.48\textwidth]{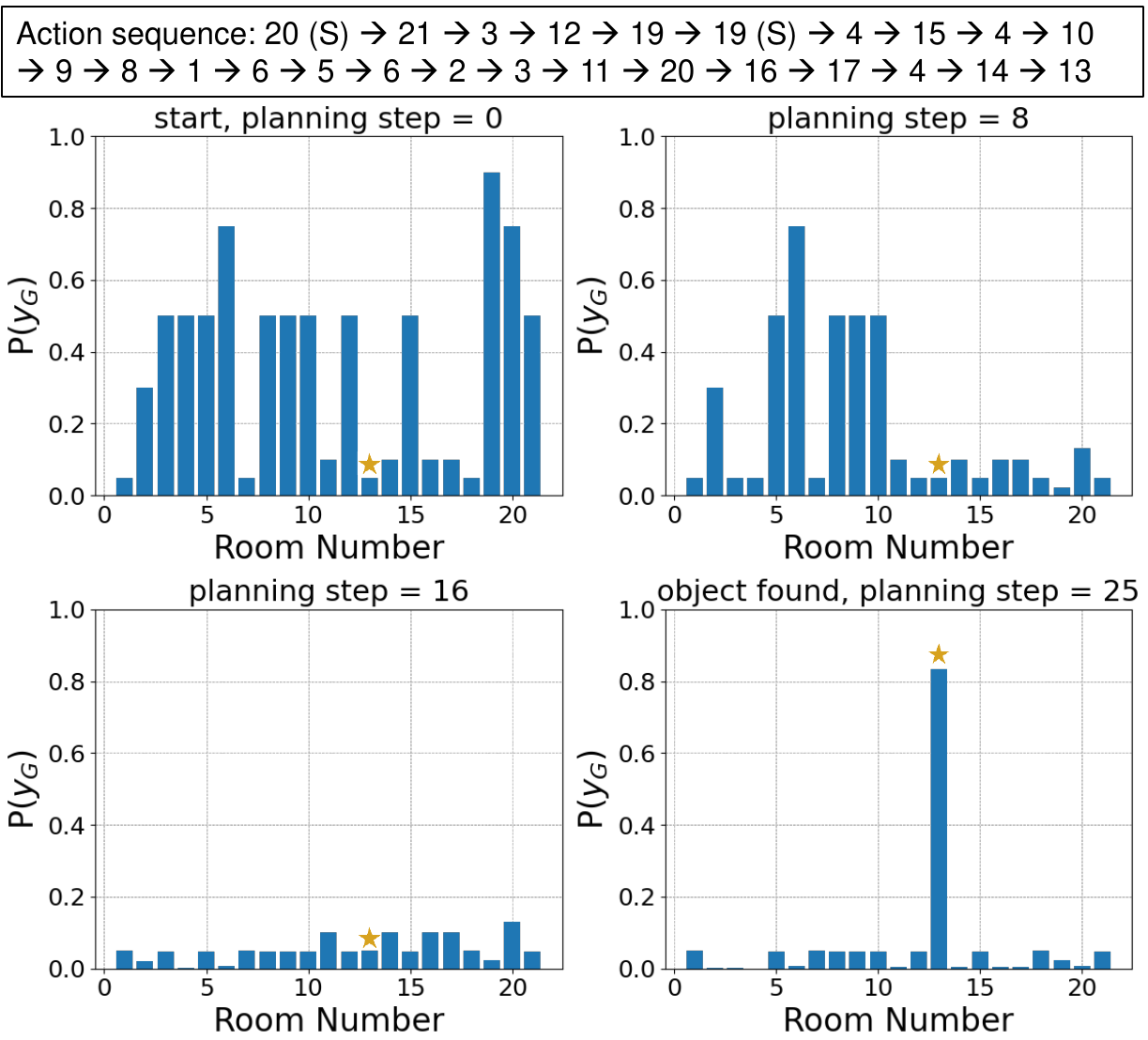}
    \caption{
    The action sequence summarizes the computed global plan on every planning step. The number represents the room number, shown in \autoref{fig:different_placements}, that the robot moves to on every time step. The number with ``(S)" represents searching the current room. The bar plot shows changes in the probability estimate of the target object as the robot moves through different rooms. The star on the bar plot marks the true object location.
    }
    \label{fig:prob_variation}
    \vspace{-7pt}
\end{figure}

The left plot in \autoref{fig:different_placements} summarizes the SPL performance on different object placements and compares the performance against the \textit{Room Coverage}. 
Finding the coffee mug in this environment is more challenging than searching static objects closely related to a specific room type, such as searching for a fire extinguisher presented in \autoref{fig:sim_quant}. 
When the object is placed in a room with a predicted probability of less than $0.5$, we observe a decrease in the SPL performance. This behavior is expected as our approach prioritizes searching rooms with higher probabilities. 
Compared to the \textit{Room Coverage}, we observe that the average SPL of SEEK is less than the \textit{Room Coverage} for object placement in a room with a probability less than $0.5$. 
The SPL of the \textit{Room Coverage} averages around 0.25 and has a larger variance than our approach. 
This result indicates that when the RSN does not have a good semantic prior of the object presence in an environment, due to the unexpected object placements or wrong prior knowledge, performing \textit{Room Coverage} can be more efficient at first until the RSN improves the prediction with object observation in the environment.

We visualize one example when the object is placed in the least expected location in the PPE storage room (Room number 13). Initially, our model predicts the probability of finding a coffee mug in the room is $0.05$. 
\autoref{fig:prob_variation} shows the sequence of the robot's global action and changes in the probability estimate of finding the object. 
At the beginning of the run, the robot moves to nearby rooms with higher probability. 
After searching rooms where a coffee mug is usually located, it visits less common rooms and finds the coffee mug after 25 global planning steps. 
After finding the target object, the updated probability of finding it in the PPE storage is much higher than in other rooms. 


\section*{C. Text Embedding Comparison for Relational Semantic Network (RSN)}
\label{app:test_embedding}
We compare the performance of the RSN using different embedding models. 
The text embedding of the object name is used as the input feature of the MLP that estimates the probability of finding the object. 
We evaluate the impact of using different embedding by evaluating the RSN prediction on a test dataset. Similar to generating the training dataset, we query GPT-4 for 40 additional object names outside the training dataset. We compare four different embedding models: a small and base version of the BGE text embedding~\cite{bge_embedding}, a text encoder of the CLIP vision-language model~\cite{radford2021learning}, and the base BERT model~\cite{devlin2019bert}. 

\autoref{tab:embedding_performance} summarizes the RSN's prediction performance and the model's size. We observe a similar prediction error between \textbf{bge-small} embedding and CLIP text embedding. The \textbf{bge-base} and \textbf{bert-base-uncased} with a larger number of model parameters perform better than smaller models. 
Qualitatively, we observe that all the models we evaluate can differentiate the room types that are more related to the target object from the other rooms.

\begin{table}[h!]
\renewcommand{\arraystretch}{1.27}
\caption{Comparison of the prediction performance of the RSN using different text embedding. The prediction error is the mean squared error on the test dataset.}
\centering
\begin{tabular}{@{}lcc@{}}
\toprule
\textbf{Model Name} & \textbf{Number of Parameters} & \textbf{Prediction Error} \\
\midrule
\textbf{bge-small}~\cite{bge_embedding} & 33 M & 0.051 \\
CLIP (Text encoder)~\cite{radford2021learning} & 65 M & 0.052 \\
\textbf{bge-base}~\cite{bge_embedding} & 109 M & 0.045 \\
\textbf{bert-base-uncased}~\cite{devlin2019bert} & 110 M & 0.043 \\
\bottomrule
\end{tabular}
\label{tab:embedding_performance}
\end{table}


\end{document}